\documentclass{article}
\usepackage[preprint]{neurips_2024}
\usepackage[utf8]{inputenc} 
\usepackage[T1]{fontenc}    
\usepackage{hyperref}       
\usepackage{url}            
\usepackage{booktabs}       
\usepackage{amsfonts}       
\usepackage{nicefrac}       
\usepackage{microtype}      
\usepackage{xcolor}         
\usepackage{graphicx}
\usepackage{amsmath, amsthm, amssymb}
\usepackage{subcaption}
\usepackage{enumitem}

\title{Token-Level Uncertainty-Aware Objective for Language Model Post-Training}

\author{%
  Tingkai Liu\quad Ari S. Benjamin\quad Anthony M. Zador \\
  Cold Spring Harbor Laboratory \\
  Cold Spring Harbor, NY, USA\\
  \texttt{\{tiliu,benjami,zador\}@cshl.edu}
}

\begin{document}
\maketitle

\begin{abstract}
  In the current work, we connect token-level uncertainty in causal language modeling to two types of training objectives: 1) masked maximum likelihood (MLE), 2) self-distillation. We show that masked MLE is effective in reducing \emph{epistemic} uncertainty, and serve as an effective token-level \textit{automatic curriculum learning} technique. However, masked MLE is prone to overfitting and requires self-distillation regularization to improve or maintain performance on out-of-distribution tasks. We demonstrate significant performance gain via the proposed training objective - combined masked MLE and self-distillation - across multiple architectures (Gemma, LLaMA, Phi) and datasets (Alpaca, ShareGPT, GSM8K), mitigating overfitting while maintaining adaptability during post-training. Our findings suggest that \textit{uncertainty-aware} training provides an effective mechanism for enhancing language model training.
\end{abstract}

\section{Introduction}
Training large language models (LLMs) through next-token prediction with a maximum likelihood objective has shown remarkable generalization capabilities in diverse tasks~\cite{chung2022scaling, ouyang2022training, touvron2023llama, wang2022self, zheng2023judging}. The power of the self-supervised training objective lies in its ability to unify a wide range of language modeling tasks (e.g. grammatical correctness, arithmetic, code generation). However, despite their power and wide adoption, LLMs still suffer from issues such as hallucinations that could stem from overfitting to the training data, particularly during the post-training stage, where the size and diversity of the training set are limited. 

In the present work, we examine the heterogeneous nature of tokens and their associated aleatoric and epistemic uncertainties more carefully \cite{hullermeier_aleatoric_2021, gupta2024languagemodelcascadestokenlevel, fadeeva2024factcheckingoutputlargelanguage}. Aleatoric uncertainty refers to the inherent irreducible stochasticity in the data, whereas epistemic uncertainty refers to model limitations that can potentially be reduced with additional information or a better model. We argue that, as opposed to the vanilla maximum likelihood estimation (MLE) objective, further performance gain could be obtained by focusing on learning tokens with high epistemic uncertainties, while avoiding overfitting by maintaining adequate aleatoric uncertainty estimation on remaining tokens. This issue is particularly prevalent in post-training, where the pre-trained model must simultaneously adapt to new response patterns (high epistemic uncertainty tokens) while retaining generalization across diverse tasks.

However, accurate uncertainty estimation requires aggregating predictions over a \emph{large} model ensemble obtained by, for example, stochastic forwards by Monte Carlo Dropout Sampling \cite{gal2016dropout} (MCDO). As MCDO could require hundreds of forward passes to converge, a simpler alternative is required to be used during training. We show that the model's predictive loss, which only requires single forward pass to compute, is a good proxy for epistemic uncertainty estimated via Bayesian Activate Learning by Disagreement (BALD) \cite{houlsby2011bayesianactivelearningclassification,kirsch2019batchbald}.

Through extensive experiments on multiple architectures (Gemma \cite{gemma,gemma2}, LLaMA\cite{touvron2023llama,touvron2023llama2,grattafiori2024llama3herdmodels}, Phi\cite{abdin2024phi3technicalreporthighly}), datasets (Alpaca, ShareGPT, GSM8K) and downstream tasks (AlpacaEval, IF-Eval, GSM8K), we show that training only on tokens with high loss (masked MLE objective) results in strong in-distribution performance gain compared to vanilla MLE baseline. As epistemic uncertainty reduces with training, the masked MLE objective also provides a natural token-level automatic curriculum learning. 

\begin{figure}[t]
    \centering
    \includegraphics[width=\linewidth]{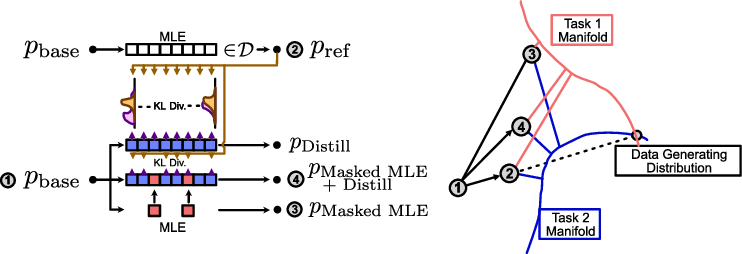}
    \caption{Proposed training procedure combines maximum likelihood with self-distillation training objective to improve both in-distribution and out-of-distribution performances.}
    \label{fig:mle_plus_distill_cartoon}
\end{figure}

Finally, we observe that training only on tokens with high epistemic uncertainty via the masked MLE objective leads to poor out-of-distribution generalization as a result of overfitting. We show that such issues could be remedied by combining masked MLE on high-loss tokens with a distillation objective on the remaining tokens as shown in Fig.~\ref{fig:mle_plus_distill_cartoon}. 

In conclusion, we propose an uncertainty-aware training objective that outperforms the predominant MLE objective in post-training across both in-distribution and out-of-distribution tasks. 

\section{Related Works}
\paragraph{Training Data Selection}
Training (core-set) data selection aims to identify a subset of the training data that can effectively represent the entire dataset. Prior arts draw inspiration from a wide range disciplines, ranging from computational geometry \cite{sener2018active}, gradient-based selection \cite{mirzasoleiman2020coresets, xu2021gradientdriven,killamsetty2021grad}, influence functions \cite{wang2020optimizing}, statistical mechanics \cite{sorscher2023neuralscalinglawsbeating} and implicit reward modeling \cite{zhou2024davirdataselectionimplicit}. In the current work, we focus on token-level data selection methods \cite{lin2025rho1tokensneed}.

\paragraph{Curriculum Learning}
Curriculum learning aims to improve model training by presenting data in a meaningful order, and has been shown to be effective across computer vision  \cite{bengio2009curriculum, weinshall2018curriculum} and language modeling \cite{elman1993learning,xu2020curriculum, zhang2018empirical} tasks. In recent years, automated methods of curriculum designs have been proposed based on model competence \cite{platanios2019competence}, model ranking \cite{sachan2016easy}, and gradient norm \cite{liu2020norm}. 

\paragraph{Uncertainty Estimation}
Uncertainty estimation \cite{hullermeier_aleatoric_2021} and related areas such as conformal prediction \cite{quach2024conformallanguagemodeling,yadkori2024mitigatingllmhallucinationsconformal}, model calibration \cite{kong2020calibrated, desai2020calibration} are foundational areas of research especially in the context of large language models \cite{malinin2021uncertainty}. Uncertainty can be estimated via Bayesian model ensembles \cite{pearce2018uncertainty} such as Monte Carlo Dropout \cite{gal2016dropout} and deep ensembles \cite{lakshminarayanan2017simple}. Uncertainty is classified into aleatoric (data/irreducible) and epistemic (model/reducible) types, which can be separately estimated either via uncertainty estimation heads \cite{nixweigend1994,kendall2017uncertainties} or Bayesian Active Learning by Disagreement \cite{houlsby2011bayesianactivelearningclassification}.

\section{Heterogeneous Token-Level Uncertainty}
Traditionally, the training loss of a single datum (i.e. single document in the training corpus) is aggregated across all tokens by taking their average (or sum) of losses. This metric is intuitive in that minimizing such metric is equivalent to maximizing the overall joint likelihood of all tokens in the entire document (we refer to this objective as \emph{vanilla MLE}), which asymptotically converges the model to the data generating distribution (as illustrated in Fig.~\ref{fig:mle_plus_distill_cartoon}\textcircled{1}).

\begin{figure}[h]
    \centering
    \includegraphics[width=.9\linewidth]{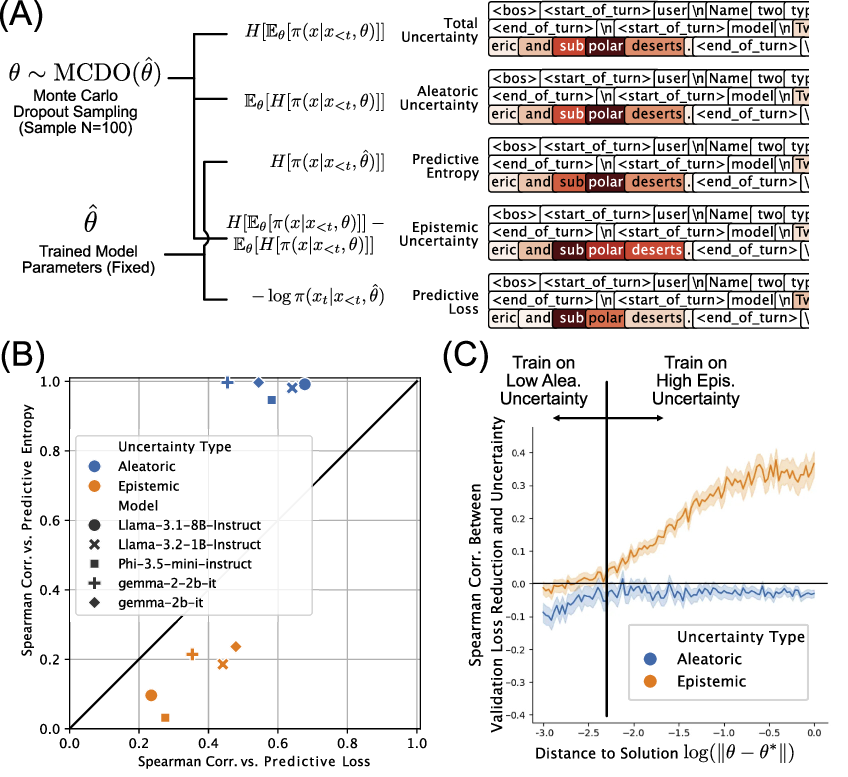}
    \caption{
        (A) Token level uncertainties, predictive loss and entropy for \texttt{Gemma-2B-it}. 
        Note that only tokens in the completion are color coded by the various uncertainty and loss metrics. 
        (B) Correlation between uncertainty (epistemic/aleatoric) and model metrics (predictive loss/entropy) in language modeling of Alpaca dataset across models.
        (C) Effect of training on different data subset with varying degree of aleatoric/epistemic uncertainty varied based on distance from solution.
    }
    \label{fig:token_uncertainty_example_gemma}
\end{figure}

However, equally weighting all tokens is not the optimal objective due to (at least) two reasons. First, when data has input-dependent noise that is not uniform across all tokens, this will affect the likelihood in a per-token and heterogeneous manner. This is a case of \emph{aleatoric} noise, referring to noise inherent to the data and due to the environment. In this case it is proper to weigh low-noise tokens higher according to the aleatoric noise level.

A second, competing force is the effect of reducible model uncertainty, called \emph{epistemic} uncertainty, which decreases with model training. If this is known, a Bayesian active learning framework prescribes that the most important data points (tokens) are those which maximally reduce model uncertainty. These can be intuitively understood as the most surprising examples (tokens). 

It is important to recognize that the appropriate weights in the MLE objective for each token differ in the aleatoric and epistemic components, and that this tradeoff changes with model training. To demonstrate this in a well-understandable example, we explored a toy linear regression model where we artificially inject known heteroscedastic aleatoric noise, and compared reduction in validation loss by training on single datum. We show that, in linear regression, training should focus on reducing epistemic uncertainty when far from optimum and focus on avoiding data with high aleatoric uncertainty when close to optimum (see Fig.~\ref{fig:token_uncertainty_example_gemma}(C) for results and Appendix.~\ref{sec:linear:alea_epis} for detailed description of the experimental procedure). This observation is confirmed by a toy MNIST classification example as shown in Appendix~\ref{sec:toy}. In the beginning of training, the epistemic uncertainty is uniform across examples, and it is preferable to prioritize low-aleatoric-noise examples. Later, the epistemic uncertainty decreases faster for low-aleatoric-noise examples, and the important points become the high-epistemic-uncertainty examples.

One of our central claims is that these effects are exaggerated in LLM training due to the nature of the task. LLM training is inherently a massively multi-task endeavor at the token level: individual tokens contribute to a wide range of linguistic and cognitive tasks. This multi-task nature of language model is made apparent when comparing the uncertainty levels at individual tokens. We observe in Fig.~\ref{fig:token_uncertainty_example_gemma}(A,B) that, across models and datasets, epistemic (resp. aleatoric) uncertainty vary greatly between tokens (see also the overall statistics of per-token losses in Fig.~\ref{fig:lm_loss_distribution} in Appendix.~\ref{appx:sec:uncertainty}). This high degree of uncertainty variability across tokens can be exploited to bias the training objective depending on whether a token is in high or low epistemic regime. 

To construct the aleatoric and epistemic uncertainty levels for LLMs, we rely on Monte Carlo dropout sampling to construct an ensemble \cite{gal2016dropout}. Define the output probability of an autoregressive model with parameters $\theta$ as $\pi(x|x_{<t},\theta)$. We can estimate the aleatoric uncertainty as the entropy of the output classes, marginalized over the ensemble:
$$U_{\text{aleatoric}}=\mathbb{E}_\theta[H[\pi(x|x_{<t})]]$$
The epistemic uncertainty can be calculated via Bayesian Active Learning by Disagreement \cite{nixweigend1994,kendall2017uncertainties}.
$$U_{\text{epistemic}}=H[\mathbb{E}_\theta[\pi(x|x_{<t})]]-\mathbb{E}_\theta[H[\pi(x|x_{<t})]]$$

While these metrics provide useful understanding about how and when tokens should be weighted, they are not practical algorithms for training as they require multiple (hundreds) forward passes to estimate. Here, we report that epistemic and aleatoric losses have definite correlations with the model's predictive loss and output entropy as shown in Fig.~\ref{fig:token_uncertainty_example_gemma}(B). In particular, the epistemic uncertainty is more correlated with predictive loss than entropy, vice versa for aleatoric uncertainty.  This means that useful approximations can be used in practice to achieve higher performance gain. 

Given these results, we propose to aggregate MLE loss only on tokens with high loss during training (masked MLE). Simultaneously, we can avoid overfitting by incorporating self-distillation which preserves information regarding aleatoric uncertainty of the remaining tokens.

\section{Experiments}
\subsection{Experimental Setup}
Given resource constraints, for our experiments on finetuning pretrained LLM using different masked objectives, we chose three smaller base models Gemma-2B, Gemma-2-2B, and Llama-3.2-1B. All models were trained using rank-32 Low Rank Adaptation (LoRA) on all linear modules with $\alpha=64$. Unless specified otherwise, all models were finetuned for 1 epoch on the training dataset with batch-size 32, at learning rate \texttt{1e-4} with cosine learning rate schedule.

\paragraph{Training \& Evaluation Datasets}
Models were trained and evaluated using a wide range of datasets/tasks as shown in Table ~\ref{tab:data}. Each trained model is evaluated on all tasks to gauge both in-distribution and out-of-distribution performances.

\begin{table}[t]
\centering
\begin{tabular}{@{}llll@{}}
\toprule
Task                  & Training & Evaluation     & Train/Test Size      \\ \midrule
Single-Turn QA        & Alpaca   & AlpacaEval 2.0 & 52K/805   \\
Multi-Turn QA         & ShareGPT & -              & 52K/-    \\
Instruction Following & -        & IF-Eval        & -/541     \\
Math. Reasoning       & GSM8K    & GSM8K          & 7.5K/1.5K \\ \bottomrule
\vspace{1pt}
\end{tabular}
\caption{
Training and Evaluation benchmarks for each task considered in the current work. 
Note that model trained using each training dataset is evaluated against all downstream tasks 
to gauge both in-distribution and out-of-distribution performances.
}
\label{tab:data}
\end{table}

The performance evaluations presented in the current work focus on comparison of training with modified objectives against the \emph{baseline} (vanilla MLE). For freeform QA tasks such as AlpacaEval, the experiment generations (models trained with objectives other than MLE) were evaluated head-to-head to the generations from the baseline (model trained with MLE), adjusted for both length and positional biases. Given the scale of the experiments, we opted to use \texttt{Qwen/Qwen2.5-7B-Instruct} model as the judge, which in Oct. 2024 was the highest ranked judge model on Judge Arena that is smaller than 50B in parameter size. For tasks with ground truth metrics such as IF-Eval and GSM8K, we report both the raw performance metrics as well as the normalized the model performance (\texttt{(model - baseline) / baseline}).

\subsection{In-Distribution Performance of Token-Level Masked MLE}

\begin{figure}[t]
    \centering
    \includegraphics[width=.8\linewidth]{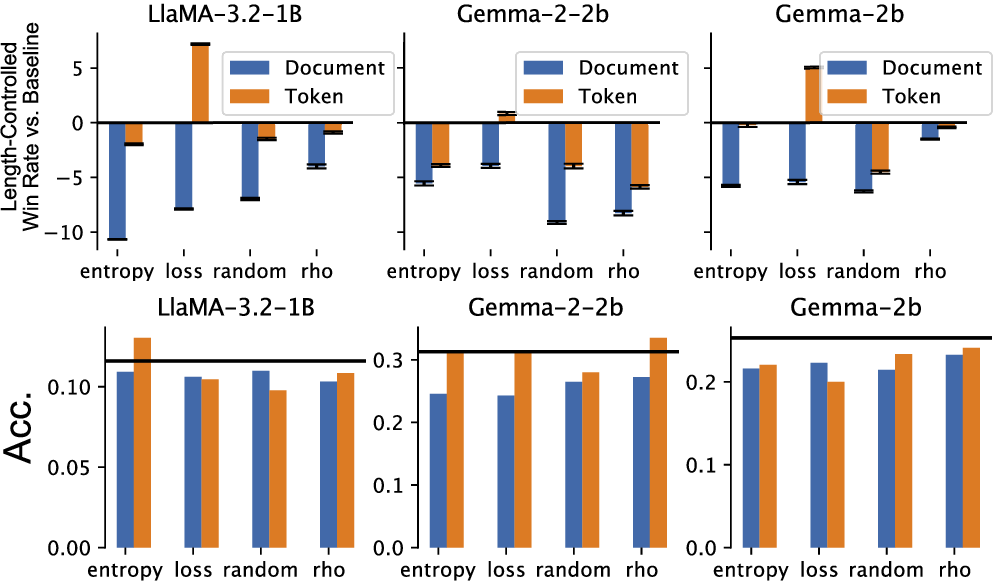}
    \caption{
        In-Distribution performance gain of token-level masked MLE compared to baseline (vanilla MLE) and document-level masked MLE. Models were trained on (top) Alpaca and (bottom) GSM8K via masked MLE objective on tokens wiht top 25\% quantile metric value.
    }
    \label{fig:in_d_perf}
\end{figure}

We show that the fine granularity of token-level masked MLE objective provides superior in-distribution performance than document-level loss masking. We choose 25\% of document/tokens to 
compute MLE losses based on either 1) entropy, 2) loss or 3) Reducible Holdout Loss (RHO). For reference, random selection was also included.

We observe that, as shown in Fig.~\ref{fig:in_d_perf} (and Fig.~\ref{fig:in_d_and_ood_perf}(A)), the finer granularity of token-level masked objective offers superior performance to the document-level counterpart. We observe that training with tokens with highest \emph{loss} is the only method that consistently outperformed baseline on AlpacaEval in a statistically significant manner. Henceforth, we refer to \emph{Masked MLE} as training on tokens with highest loss via the MLE objective.

\begin{figure}[t]
    \centering
    \includegraphics[width=\linewidth]{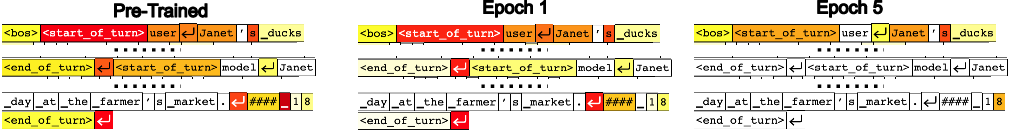}
    \caption{
        Example of automatic curriculum learning as a result of training on high epistemic uncertainty (color-coded) tokens. Note that while both prompt and response tokens are color coded, losses are only computed and propagated for response tokens during training.
    }
    \label{fig:gsm8k_acl_example}
\end{figure}

As training progresses, the distribution of epistemic uncertainty naturally progresses from structural patterns (e.g., ``<end\_of\_turn>'') to complex content (e.g., arithmetic) as shown in Fig.~\ref{fig:gsm8k_acl_example}. Thus providing a new approach to automatic curriculum at a token-level, allowing for fine-grained control over the training process.

\subsection{Regularization via Self-Distillation Improves OOD Performance}

\begin{figure}[t]
    \centering
    \includegraphics[width=.6\linewidth]{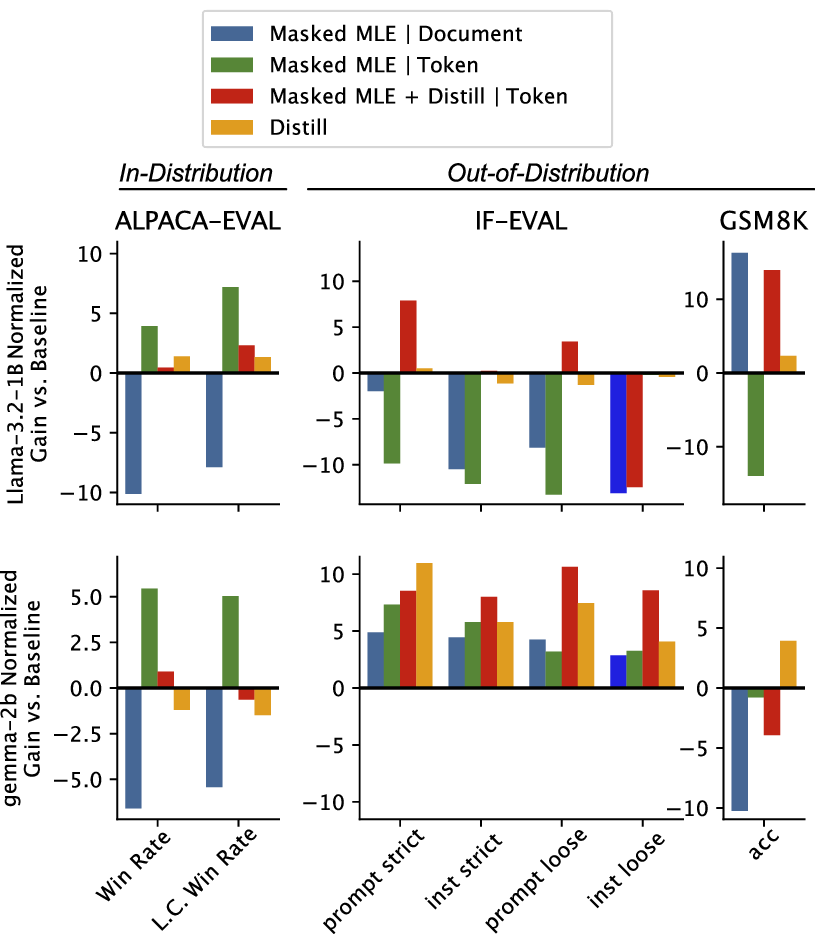}
    \caption{Downstream performance of (top) Llama-3.2-1B and (bottom) Gemma-2B trained on Alpaca dataset with different training objectives.}
    \label{fig:all_perf}
\end{figure}

\begin{figure}[t]
    \centering
    \includegraphics[width=.8\linewidth]{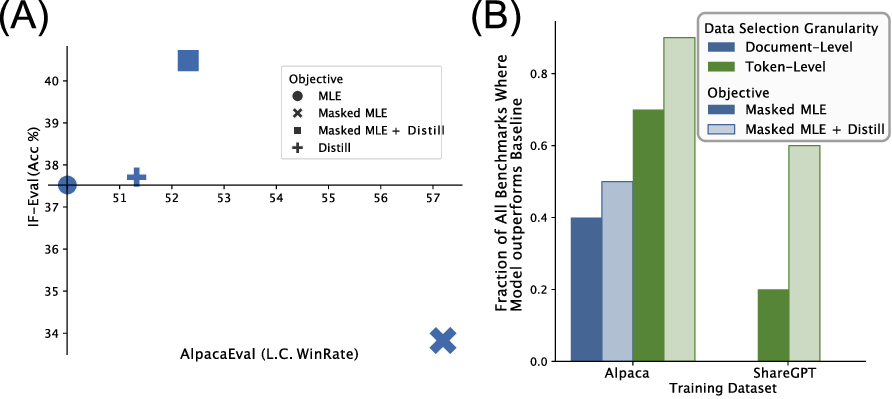}
    \caption{(A) Single-Turn QA (in-distribution) and Instruction-Following (OOD) performance gain over MLE baseline.
    (B) Fraction of all tasks where a training objective beating baseline.
    }
    \label{fig:in_d_and_ood_perf}
\end{figure}

The improved performance of masked MLE comes at the cost of overfitting to the training dataset. As shown in Fig.~\ref{fig:all_perf} and Fig.~\ref{fig:in_d_and_ood_perf}(A), the improved Alpaca-Eval performances from finetuning Llama-3.2-1B and Gemma-2B on Alpaca dataset via masked MLE objective result in deterioration of both IF-Eval and other downstream task performances.

To address the issue of over-fitting, we propose to incorporate self-distillation to ensure aleatoric uncertainty is captured by the model on tokens with low epistemic uncertainty.
Thus resulting in the final training objective of \textit{Masked MLE + Distill} as shown in Fig.~\ref{fig:mle_plus_distill_cartoon} and Fig.~\ref{fig:in_d_and_ood_perf}, given as:
\begin{equation*}
\begin{split}
&\mathcal{L}^i_t = \begin{cases}
    - \log p_\theta(x^i_t|x^i<t) & \text{high loss} \\
    \sum_{x\in \mathcal{V}} - p_\text{ref}(x|x^i_{<t})\log p_\theta(x|x^i<t) & \text{otherwise,} \\
\end{cases}    
\end{split}
\end{equation*}
where $p_\text{ref}$ is the base model finetuned on the same dataset via the vanilla MLE objective.

As shown in Fig.~\ref{fig:all_perf}, this training objective simultaneously improves in-distribution performance and out-of-distribution generalization for both Llama-3.1-1B and Gemma-2B model trained on Alpaca and ShareGPT (see Fig.~\ref{fig:in_d_and_ood_perf} (right)) datasets. 

We note that the 25\% quantile was chosen arbitrarily and tuning of this hyperparameter could potentially lead to stronger overall performance, which we leave for furture studies.

\section{Limitations}
\paragraph{Data Selection Granularity and Computational Savings}
While token-level selection provides fine-grained control over training, its computational benefits are more limited compared to document-level selection due to the underlying mechanics of transformer computation. Token-level selection requires the full model forward and backward, with savings only in the final classification layer which is negligible for very large models.

This modest computational benefit of token-level reselection suggests its primary value lies in its ability to induce better learning dynamics and approximate reward signals, rather than in training efficiency. Future architectures that allow for more efficient sparse attention computation could potentially improve these savings, but with current transformer implementations, document-level selection remains substantially more efficient for reducing computational costs.

\paragraph{Data Curriculum and Model Curriculum}
While our token-level selection method demonstrates the emergence of an effective data curriculum, the model capacity does not change commensurately. A more comprehensive curriculum would jointly optimize both the data distribution and model complexity, and we leave the exploration of the Pareto frontier of data and model curricula for future works.

\section{Conclusion}
In conclusion, we show that uncertainty estimation provides a new way of examining training objective in language model at a token-level. We propose a novel training objective that combines masked maximum likelihood and distillation objective that improve model performance on in-distribution and out-of-distribution downstream tasks.

\clearpage
\bibliographystyle{plainnat}
\bibliography{custom}

\clearpage
\newpage
\appendix


\appendix

\section{Token-Level Uncertainty via Dropout Ensemble}\label{appx:sec:uncertainty}
\begin{figure}[t]
    \centering
    \includegraphics[width=.6\linewidth]{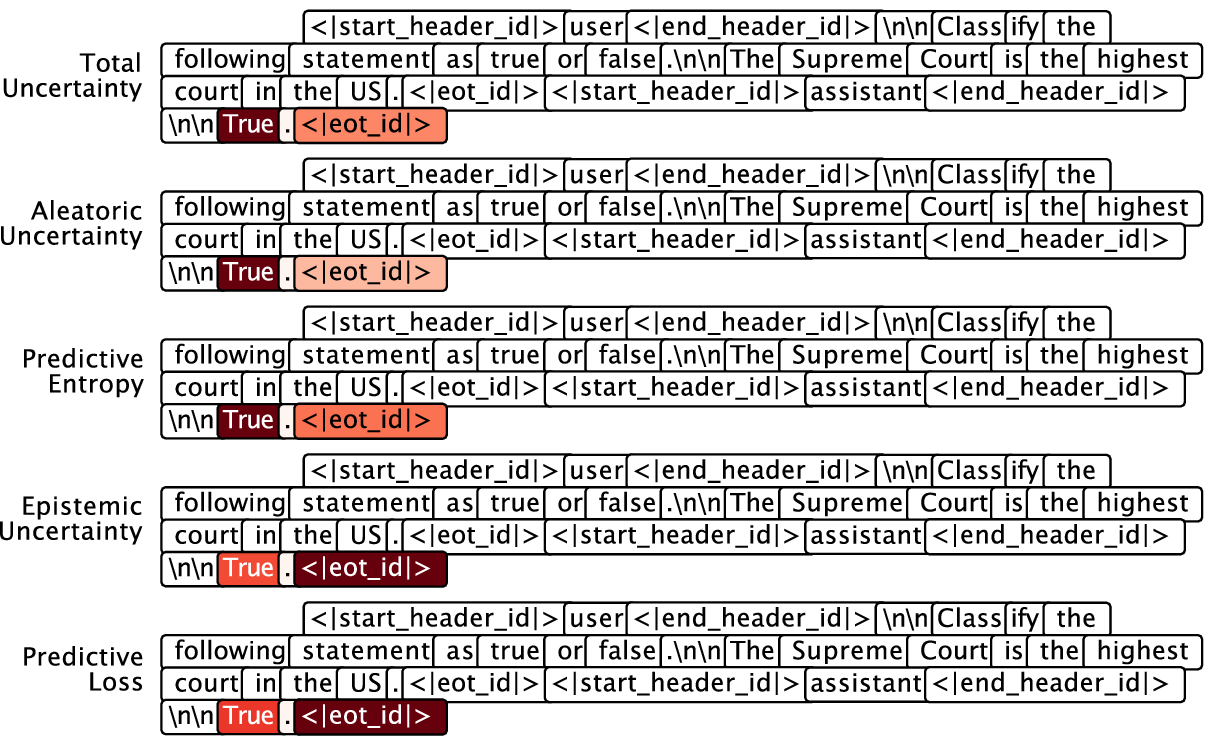}
    \caption{Token level uncertainty and loss metrics for \texttt{Llama-3.2-8B-Instruct}.}
    \label{fig:token_uncertainty_example_llama}
\end{figure}

\begin{figure}[t]
    \centering
    \centering
    \includegraphics[width=.8\linewidth]{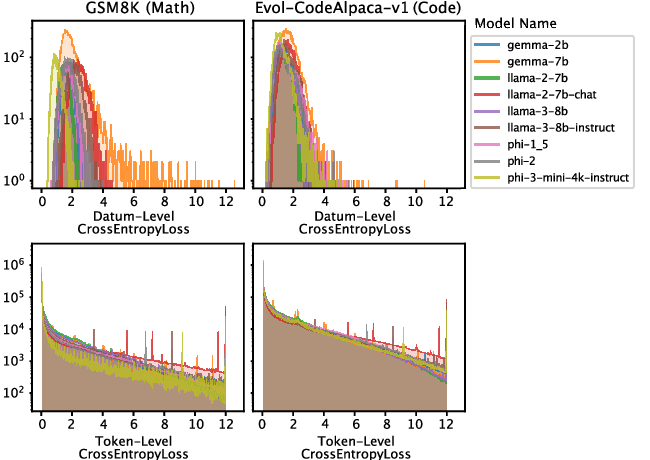}
    \caption{Distribution of datum-level and token-level losses across models and datasets.}
    \label{fig:lm_loss_distribution}
\end{figure}

LLM training is inherently a multi-task as individual tokens contribute to a wide range of linguistic and cognitive tasks, such as:
\begin{itemize}[leftmargin=*]
    \item Sentiment analysis (e.g., ``The movie was \boxed{excellent}!'')
    \item Grammar and syntax (e.g., ``The cat sat \boxed{on} the mat.'')
    \item Variable definition in code generation (e.g., ``let \boxed{x} = 5;'')
    \item Arithmetic computation (e.g., ``7 * 6 = \boxed{42}'')
\end{itemize}
Which is demonstrated by both the high degree of loss variability at the token-level as shown in Fig.~\ref{fig:lm_loss_distribution}, as well as the example in Fig.~\ref{fig:token_uncertainty_example_llama}.

\begin{figure*}[t]
    \centering
    \centering
    \includegraphics[width=\linewidth]{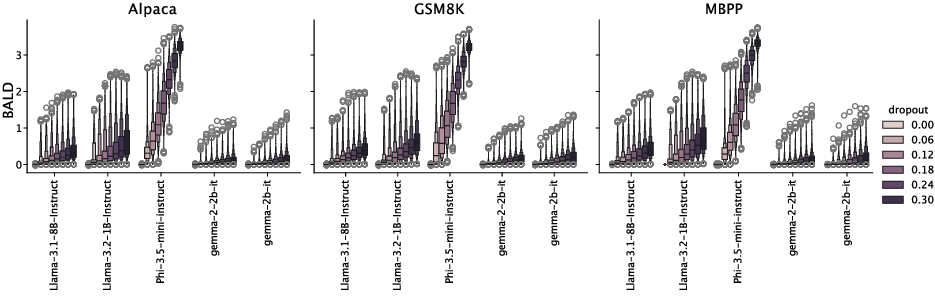}
    \caption{
        Distribution of BALD metric across dropout rate for different models.
        Note that BALD scales monotonically with increasing parameter noise, consistent with
        the notion of epistemic uncertainty. Also note that different models have different 
        degree of sensitivity to dropout ratio, partly due to the difference in the number of
        dropout modules in the model. For example \texttt{Phi-3.5-mini-instruct} have 3 dropout modules at various levels of the model whereas LlaMA models only have attention dropout.
    }
    \label{fig:bald_dist}
\end{figure*}

Here token-level uncertainties are estimated by Monte Carlo Dropout Sampling (MCDO) with 100 samples at dropout rate of 0.1. The total uncertainty is given by the entropy of the empirically averaged predictive distribution, aleatoric uncertainty is given by the empirical average of entropy of each MCDO sampled predictive distribution and epistemic uncertainty is their difference. In effect, the epistemic uncertainty is equivalent to the Bayesian Active Learning by Disagreement (BALD) objective. We note that measures of epistemic uncertainty should increase monotically with parameter noise (dropout rate), which is confirmed for the BALD objective as shown in Fig.~\ref{fig:bald_dist}.

\begin{figure}[t]
    \centering
    \includegraphics[width=.8\linewidth]{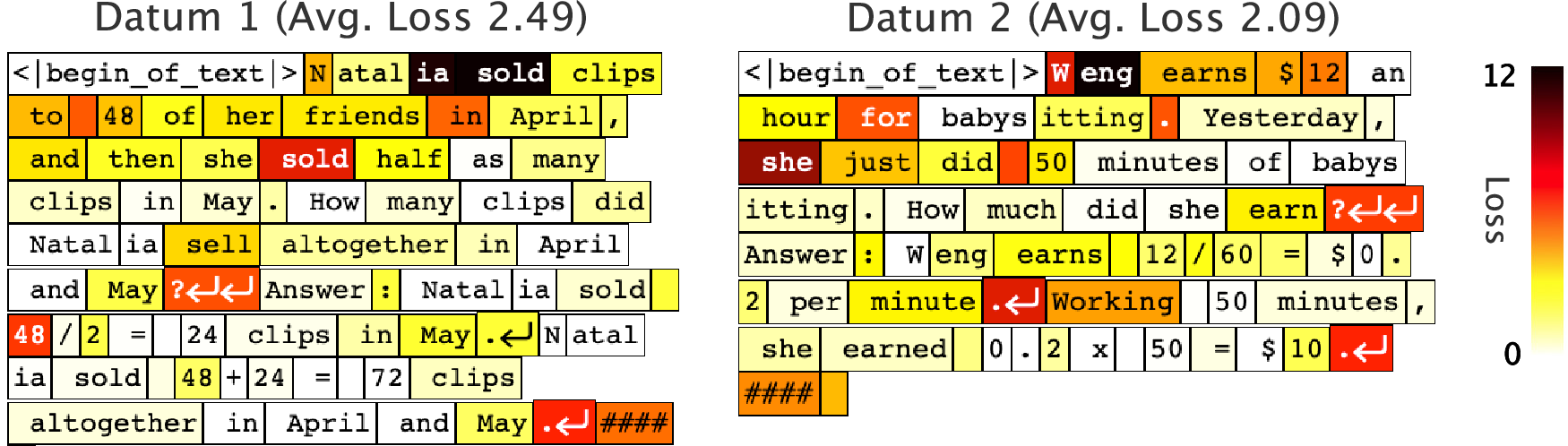}
    \caption{Token-level and datum-level losses of Phi-3.5-Mini-Instruct on two data points in the GSM8K dataset.}
    \label{fig:token_loss_example}
\end{figure}

\section{Weighted Classification in Sequential Generative Modeling}\label{sec:weighted_class}
In this section we formulate the weighted classification objective which unifies a wide range of training objectives as well as data selection methods in sequential modeling.

Consider the following objective
\begin{equation}
    \max_{\theta}\;\;\sum_{(i,t)}\sum_{x \in \mathcal{V}}w_{i,t}(x)\;\;\log p_\theta\bigl(x | x^i_{<t}\bigr)
    \label{eqn:weighted_seq}
\end{equation}
where $x^i_{<t}$ is the context (the tokens before position $t$ in sequence $i$), $x \in \mathcal{V}$ ranges over the \emph{vocabulary} $\mathcal{V}$ of size $|\mathcal{V}| = V$ and $w_{i,t}(x)$ is a token-level \emph{weight} that scales the log-likelihood of predicting token $x$ given context $x^i_{<t}$.

The formulation in \eqref{eqn:weighted_seq} is general by design, admits a wide range of training objectives as special cases:
\begin{itemize}[leftmargin=*]
    \item {\bf (Masked) Maximum Log-Likelihood} Equivalent to CrossEntropyLoss with 1-hot labels, the weights are given as
        \[w_{i,t}(x) = \delta(x=x^i_t),\]
    which is the standard training objective in pre-/post-training (supervised fine-tuning) of language models.
    This formulation further admits masked MLE objective which encompasses training with core-set selection, where weights are given as 
    \[ w_{i,t}(x) = \delta(x=x^i_t) m^i_t\]
    for some token-level mask $m^i_t \in \{0,1\}$ that either includes or discards a token from the training objective.
    \item {\bf (Forward) Knowledge Distillation} Equivalent to CrossEntropyLoss with full teacher distributions, where the weights are given as.
        \[w_{i,t}(x)\;=\;p_{\text{teacher}}(x|x^i_{<t}),\]
        This formulation further admits a special case of truncated KL divergence, 
        \[w_{i,t}(x)\;=\;p_\text{teacher}(x|x^i_{<t}) ~\delta\bigl(x \in X^i_t),\]
        where $X^i_t, |X^i_t| \leq |\mathcal{V}|$ is some set of admissible tokens. Truncated KL divergence has previously been been shown to improve student LM performance performance. Note that the this formulation does not include \emph{reverse} distillation which minimizes the objective $D_{KL}(p_\text{teacher} || p_\text{student})$ (which includes an entropy term of the student model in the objective). However, since the effectiveness of reverse distillation is unclear, we omit this training objective from our formulation.
    \item {\bf Policy Gradient} For techniques such as Proximal Policy Optimization (PPO) and Group Relative Policy Optimization (GRPO), the weights are given as
        \[w_{i,t}(x)\;=\;A(x | x^i_{<t})\]
        where $A(\cdot)$ is the advantage function of token $x$ given context. Note that we omit the KL regularization terms here for brevity.
    \item {\bf Weighing due to aleatoric (heteroskedastic) noise} When labels are noisy and furthermore this noise on labels is input-dependent, and not uniform across examples, the likelihood function changes. The effective result of this is a per-example weighting of the log-likelihood. In Least-Squares regression, for example, adjusting for the input-dependence variance of noise results in Weighted Least Squares. In a classification setting, one instead considers the \textit{oracle} distribution $p_\text{oracle}$ as capturing the true aleatoric label noise. If this is known (which is rare, although such datasets are becoming more common, e.g. \cite{wei2021learning}), then, 
    \[w_{i,t}(x)\;=\;p_{\text{oracle}}(x|x^i_{<t}).\]
    Note the similarity of this objective to model distillation.
    \item {\bf Active learning via epistemic uncertainty}  In an active learning setting, the goal is to select examples for optimal learning, considering one's current uncertainty. A principled formulation of this is Bayesian active learning, a task-agnostic formulation which aims to decrease the uncertainty over the model parameters. This can be expressed through a data masking function $\delta(x=x^i_t)$ that selects points which maximize expected information gain:
    \[\mathbb{E}_{x \sim p_\theta(\cdot|x^i_{<t})}\left[D_{KL}(p(\theta|x^i_{<t},x) | p(\theta|x^i_{<t}))\right],\]
    where $p(\theta|x^i_{<t})$ represents the current posterior over model parameters and $p(\theta|x^i_{<t},x)$ is the updated posterior after observing example $x$. In practice, this expectation can be approximated using ensemble methods or Monte Carlo dropout, leading to computationally tractable uncertainty estimates that guide the selection of informative examples for training (e.g. \cite{kirsch2019batchbald}).
\end{itemize}

\subsection{Comparing Three Special Cases}
Assuming we are given training labels $\{x^i_t\}_{i,t}$ sampled from some oracle distribution $p_{teacher}$, and assume that there exists an oracle reward model (and consequently an oracle advantage function $A$), we argue that the performance of the student model trained via the three above objectives would follow the order of:
\[
    \text{Policy Gradient} \geq \underbrace{\text{Knowledge Distillation} \geq \text{MLE}}_{\text{Distribution Matching}}
\]

\paragraph{Policy Gradient vs. Distribution Matching - Task Alignment}
It is easy to see that Policy Gradient and distribution matching objectives are \emph{equivalent} when $A(x|x^i_{<t}) = p_\text{teacher}(x|x^i_{<t})$. This corresponds to perfect ``task alignment'', where the teacher model is optimized for the downstream task of interest. This assumption holds true for tasks such as image classification, as evidenced by the high correlation ($> 0.95$) between model likelihood scores and classification accuracy, and by the asymptotic convergence of maximum likelihood training towards optimal performance (e.g. super-human results on benchmarks like ImageNet).

However, for language modeling, the alignment between downstream task performance and distribution matching to the data generating distribution varies substantially across domains. For example, open domain QA tasks would exhibit higher correlation between likelihood and human preferences as compared to structured reasoning tasks such as maths and code generation.

As shown by DeepSeek-R1, pure reward maximization via policy gradient can lead to state-of-the-art performance on downstream tasks with oracle reward models (e.g. reasoning tasks with objective ground truths). However, for majority of open domain tasks that do not have ground truth reward, training with RL is prone to reward hacking and requires frequent data labeling and retraining of reward model to ensure that the reward model adapts to the changing distribution of the underlying language model. Fortunately, for such tasks, results above shows that distance to data generating distribution is predictive of task performance. 

\section{Data and Model Curriculum in Toy Models}\label{sec:toy}
In this section, we study the effect of biasing training on different samples based on model performance and uncertainty estimations in two toy modes: linear regression and MNIST classification with MLP.

\subsection{Linear Regression}
Consider the problem of linear regression, where for a linear system of equation $Zw = d$, we want to find the optimal 
parameter $w$ by solving the following optimization problem:
\begin{equation}
    \min_{w} \mathcal{L}(w) = \min_w \frac{1}{2}\|Zw - d\|_2^2
\end{equation}
where for $Z \in \mathbb{R}^{N\times D}$ is the measurement matrix,
$w \in \mathbb{R}^D$ is parameters and $d \in \mathbb{R}^N$ is the observation.
Note that the linear regression problem can be applied to a wide range of problems including both polynomial regression 
and sequence modeling (e.g. autoregressive modeling).

The linear optimization problem can be solved via gradient descent where the gradient update is given as 
\begin{equation}
    w^{k+1} = w^k - \lambda \cdot \nabla_w \mathcal{L}(w) = w^k - \lambda \cdot Z^T\left(Zw - d\right)
\end{equation}

\begin{figure}[t]
    \centering
    \begin{subfigure}[b]{\linewidth}
        \centering
        \includegraphics[width=.5\linewidth]{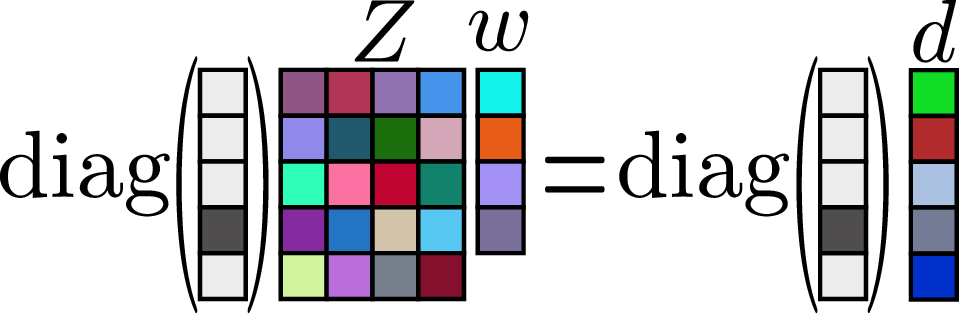}
        \caption{Training on data subset}
        \label{fig:linear_system_data}
    \end{subfigure}
    \begin{subfigure}[b]{\linewidth}
        \centering
        \includegraphics[width=0.7\linewidth]{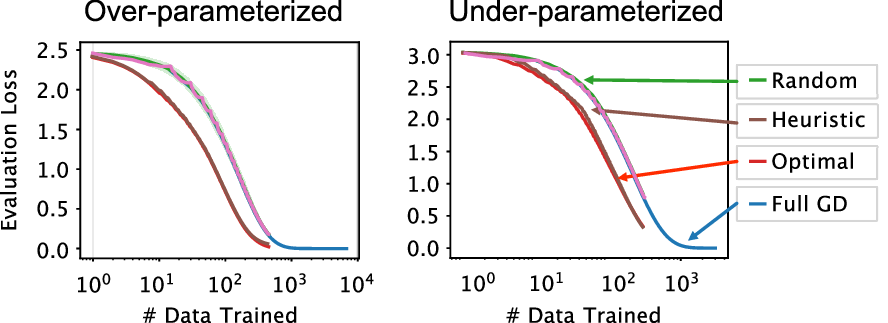}
        \caption{Results of Data Subset}
        \label{fig:results_linear_data}
    \end{subfigure}

    \caption{Results of linear regression with training on data subset .}
\label{fig:results_linear_system_all}
\end{figure}

Here, we consider Data Subset Selection (Fig.~\ref{fig:linear_system_data}), which is akin to masked MLE objective discussed in the current work. Given a indicator function of the data entry $\mathbb{I}_\text{D}\in \{0,1\}^N$, the updated linear system and the associated gradient update is given as \begin{equation}
    \begin{split}
        \text{Diag}(\mathbb{I}_\text{D}) \cdot Z w = d \\
        \nabla_w\mathcal{L}(w) \leftarrow \left(\text{Diag}(\mathbb{I}_\text{D}) Z\right)^T \left(Zw - d\right)
    \end{split}
\end{equation} 
We formulate the problem of finding the optimal subset $\mathbb{I}_\text{D}$ in a greedy fashion: for each training step given current estimate of the parameters $w^k$, find the optimal subset such that the training loss $\mathcal{L}(w^{k+1})$ is minimized. 

For finding the optimal data subset $\mathbb{I}_\text{D}$, it can be shown that the optimal $\mathbb{I}_\text{D}$ can be written as:
\begin{equation}
    \min_{\mathbb{I}_\text{D}} \mathcal{L}(w^{k+1}) = \min_{\mathbb{I}_\text{D}} \frac{1}{2}\|(I - \lambda ZZ^T \mathbb{I}_\text{D})\underbrace{(Zw^k-y)}_{\epsilon^k}\|_2^2
\end{equation}
where $w^k$ is the parameter estimate at training step $k$, $\epsilon^k = Zw^k-y \in \mathbb{R}^N$ is the corresponding residual error for the training dataset. It is easy to see that the optimal choice of $\mathbb{I}_\text{D}$ is dependent both on the current residual error and the covariance of training data $ZZ^T$. If the training data is uncorrelated (or whiten-ed), which results in the covariance matrix $ZZ^T$ being an identity matrix, then the optimal choice of $\mathbb{I}_\text{D}$ is exactly $\text{Diag}(\text{Top-K}(\epsilon^k))$, where the selected data are those that have the largest current residual error.

While the results above are only exact if very strong assumptions are place on the data matrix $Z$, we found that, in practice, these heuristics remain highly effective for arbitrary $Z$. In Fig.~\ref{fig:results_linear_system_all}, we compared the evaluation loss of models trained via the two heuristics on hold-out data against the vanilla gradient descent using all of model parameters/gradients and all of the dataset (Full GD). We also performed exhaustive search to find the globally optimal solution of $\mathbb{I}_\text{D}$ at each step. Finally, we also include random subset selection as baselines. 

As expected, we note that the globally optimal solution (shown in red) that converges to the optimal solution faster than the vanilla gradient descent algorithm for a fixed compute (in terms of number of data trained or number of parameters updated).

In conclusion, we show that data subset selection during training can be done effectively and efficiently via the maximum error heuristics for linear systems.

\subsubsection{Effective of Data Uncertainty on Training}\label{sec:linear:alea_epis}
To measure the effect of training on data with different degree of epistemic and aleatoric uncertainties, we first created a 5th degree polynomial with random coefficient as ground truth, and sampled 20 $(x,y)$ pairs from the ground truth polynomial as training data. 
100 uniformly spaced data in range $x\in[-1,1]$ and their corresponding $y$ values are computed as held-out validation data.

Aleatoric uncertainty is controlled by adding heteroscedastic Gaussian noise with known standard deviation ($0.1\cdot\text{Unif}(0,1)$) to the $y$ value of each of the training datum. Epistemic uncertainty is estimated by first adding 0-meaned Gaussian noise with standard deviation 0.002 to the coefficients of the polynomial, and computing the the variance of 1000 $y$ predictions for each $x$ (with 1000 samples of the Gaussian noise). 

We simulate different stages of training by adding 0-meaned Gaussian with standard deviation in range $[10^{-3}, 1]$ to the ground truth coefficients. For each noisy coefficient value, we perform 1 gradient descent step with learning rate 0.01 on each of the 20 data points (with aleatoric noise added), and compute the amount of validation error \emph{decrement} before and after training. This allows us to have accurate per-datum information on the impact of training on a given datum, which we can relate to the amount of aleatoric and epistemic uncertainty of the said datum. 

To evaluate how training on data with varying degree of aleatoric/epistemic uncertainty affects model performance at different stages of training, we computed Spearman Ranked correlation between uncertainty level and validation error decrement as shown in Fig.~\ref{fig:token_uncertainty_example_gemma}(C). We observe that when coefficients are far from ground truth, epistemic uncertainty is important for improving model performance. In contrast, as model converges to ground truth, training on data with high epistemic uncertainty can lead to worsening in model performance, and we instead ought to focus on avoiding overfitting to data with high aleatoric uncertainty.

\subsection{MNIST Classification with MLP}
We applied the maximum-error data selection heuristic to the more complex problem of MNIST Classification with MLP model. Note that this problem differs from that of the linear regression problem described in the previous section both in terms of its complexity but also the optimization's convexity, as image classification via MLP is known to be a highly non-convex optimization problem. 

We chose a MLP model with two hidden layers of sizes 32 and 16 respectively and \texttt{ReLU} activation function. The model is trained on the MNIST training set for 20 epochs using a constant learning rate of 0.01 and batch-size of 256 with Stochastic Gradient Descent (SGD) optimizer. 

\begin{figure}[t]
    \centering
    \centering
    \includegraphics[width=.6\linewidth]{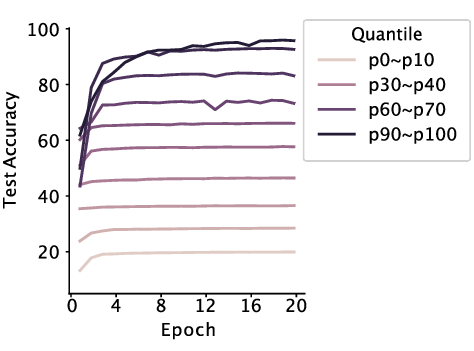}
    \caption{Test accuracy vs. training epoch when trained on data in different loss quantile.}
    \label{fig:mnist_final}
\end{figure}

To validate the effectiveness of the heuristics, we compared the test accuracy of the model trained using data with losses in different quantiles. Note that, as opposed to linear regression, the larger number of training data and parameters makes computing the quantiles for data losses globally (across all training data) infeasible. Instead, we opted to computed the quantiles \emph{in-batch}. In Fig.~\ref{fig:mnist_final}, we show the test accuracy across training epochs for training on 10\% subsets of loss. We observe that test accuracy increases almost monotonically with increasing quantiles, suggesting that the heuristics of choosing the data that require the most amount of update remains effective for the non-convex MNIST problem.

\end{document}